\documentclass{article}
\usepackage{spconf,amsmath,graphicx}

\usepackage{multirow,comment,color}
\usepackage{algorithm}
\usepackage{algpseudocode}
\usepackage{balance}
\usepackage{enumerate}
\usepackage{url}
\usepackage{epstopdf}
\usepackage[inline]{enumitem}
\usepackage[sort,nocompress]{cite}

\def\reg{{\rm\ooalign{\hfil
     \raise.07ex\hbox{\scriptsize R}\hfil\crcr\mathhexbox20D}}}

\title{Multilingual Bottleneck Features for Query by Example Spoken Term Detection}
\name{\em Dhananjay Ram, Lesly Miculicich, Herv\'e Bourlard\thanks{The research is funded by the Swiss NSF project `PHASER-QUAD', grant agreement number 200020-169398.}}
\address{Idiap Research Institute, Martigny, Switzerland \\
 \'Ecole Polytechnique F\'ed\'erale de Lausanne (EPFL), Switzerland}
\begin{document}
%
\maketitle
\begin{abstract}
State of the art solutions to query by example spoken term detection (QbE-STD) usually rely on bottleneck feature representation of the query and audio document to perform dynamic time warping (DTW) based template matching. Here, we present a study on QbE-STD performance using several monolingual as well as multilingual bottleneck features extracted from feed forward networks. Then, we propose to employ residual networks (ResNet) to estimate the bottleneck features and show significant improvements over the corresponding feed forward network based features. The neural networks are trained on GlobalPhone corpus and QbE-STD experiments are performed on a very challenging QUESST 2014 database.
\end{abstract}

\section{Introduction}\label{intro}
Query-by-example spoken term detection (QbE-STD) is the task of detecting audio documents from an archive, which contain a spoken query provided by a user. In contrast to textual queries in keyword spotting, QbE-STD requires spoken queries which enables a language independant search without the need of a full speech recognition system. The search is performed in the acoustic feature domain without any language specific resources, making it a zero-resource task.

The QbE-STD systems primarily involve the following two steps: (i) extract acoustic feature vectors from both the query and the audio document and (ii) employ those features to compute the likelihood of the query occurring somewhere in the audio document as a sub-sequence. Different types of acoustic features have been used for this task: spectral features~\cite{park2008unsupervised, chan2013model}, posterior features (posterior probability vector for phone or phone-like units)~\cite{zhang2009unsupervised, rodriguez2014high} as well as bottleneck features~\cite{szoke2015copingwith, chen2016unsupervised}. 
The matching likelihood is generally obtained by computing a frame-level similarity matrix between the query and each audio document using the corresponding feature vectors and employing a dynamic time warping (DTW)~\cite{rodriguez2014high, szoke2015copingwith} or convolutional neural network (CNN) based matching technique~\cite{ram2018cnn}. Several variants of DTW have been used: Segmental DTW~\cite{zhang2009unsupervised, park2008unsupervised}, Slope-constrained DTW~\cite{hazen2009query}, Sub-sequence DTW~\cite{muller2007information}, Subspace-regularized DTW~\cite{Ram_IEEETASLP_2018, ram2017subspace} etc. State of the art performance has been achieved using bottleneck features with DTW~\cite{szoke2015copingwith}.

Bottleneck features~\cite{hinton2006reducing, yu2011improved, vesely2012language} are low-dimensional representation of data generally obtained from a hidden bottleneck layer of a feed forward network (FFN). This bottleneck layer has a smaller number of hidden units compared to the size of other layers. The smaller sized layer constrains information flow through the network which enables it to focus on the information that is necessary to optimize the final objective. Bottleneck features have been commonly estimated from auto-encoders~\cite{hinton2006reducing} as well as FFNs for classification~\cite{yu2011improved}. Language independent bottleneck features can be obtained using multilingual objective function~\cite{vesely2012language}.

In this work, we present a performance analysis of different types of bottleneck features for QbE-STD. For this purpose, we train FFNs for phone classification using five languages to estimate five distinct monolingual bottleneck features. We also train multilingual FFNs using multitask learning principle~\cite{caruana1997multitask} in order to obtain language independent features. We used a combination of three and five languages to analyze the effect of increasing the language variation for training.

Previous studies have shown the effectiveness of convolutional neural network (CNN) for acoustic modeling in speech recognition~\cite{abdel2014convolutional, sainath2015convolutional}. Residual networks (ResNet) is a special kind of CNN which is effective for learning deeper architectures and has been shown to be very successful for image classification~\cite{he2016deep} as well as speech recognition~\cite{xiong2016achieving, zhang2017very}. This inspired us to use ResNets instead of FFNs to estimate monolingual and multilingual bottleneck features for QbE-STD. To the best of our knowledge, this is the first attempt to use ResNets for bottleneck features estimation.



In the rest of the paper, we present the multitask learning approach used to train the multilingual networks in Section~\ref{sec:multitask}. Then, we explain the monolingual and multilingual architectures using FFNs and ResNets in Sections~\ref{sec:ffn} and ~\ref{sec:resnet} respectively. Later, we describe the experimental setup in Section~\ref{sec:exp-setup}, and we evaluate and analyze the performance of our models using QUESST 2014 database in Section~\ref{sec:exp}. Finally, we present our conclusions in Section~\ref{sec:con}.
 
\section{Multitask Learning} \label{sec:multitask}
Multitask learning~\cite{caruana1997multitask, vesely2012language} have been used to exploit similarities across tasks resulting in an improved learning efficiency when compared to training each task separately. Generally, the network architecture consists of a shared part and several task-dependent parts.
In order to obtain multilingual bottleneck features we model phone classification for each language as different tasks, thus we have a language independent part and a language dependent part. The language independent part is composed of the first layers of the network which are shared by all languages forcing the network to learn common characteristics. The language dependent part is modeled by the output layers (marked in red in Figures~\ref{fig:ffn} and \ref{fig:resnet}), and enables the network to learn particular characteristics of each language. In the following sections we present different architectures that we use to obtain the multilingual bottleneck features as well as monolingual ones for comparison.

\section{Feed Forward Networks}\label{sec:ffn}
\begin{figure}
 \centering
 \centerline{\includegraphics[width=\linewidth]{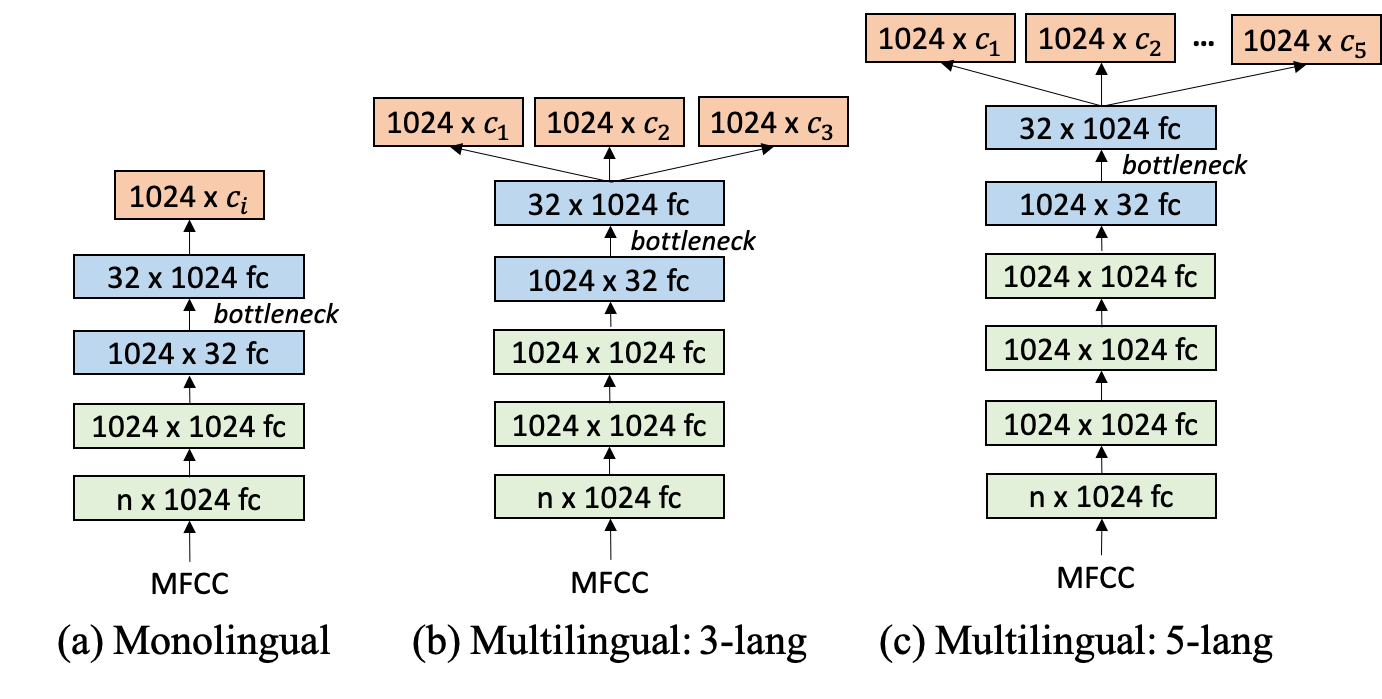}}
 \caption{Monolingual and multilingual feed forward network architectures for extracting bottleneck features using multiple languages. $c_i$ is the number of classes for the $i$-th language and $n$ is the size of input vector.}
 \label{fig:ffn}
\end{figure}

Feed forward networks have been traditionally used to obtain bottleneck features for speech related tasks~\cite{yu2011improved, vesely2012language, szoke2015copingwith}.  Here, we describe the different architectures employed in this study as shown in Figure~\ref{fig:ffn}: 
\begin{enumerate}[label=(\alph*)]
 \item Monolingual: our monolingual FFN architecture, consists of 3 fully connected layers of 1024 neurons each, followed by a linear bottleneck layer of 32 neurons, and a fully connected layer of 1024 neurons. The final layer feeds to the output layer of size $c_i$ corresponding to number of classes (e.g. phones) of the $i$-th language.
 \item Multilingual (3 languages): this architecture consists of 4 fully connected layers having 1024 neurons each, followed by a linear bottleneck layer of 32 neurons. Then, a fully connected layer of 1024 neurons feeds to 3 output layers corresponding to the different training languages. The 3 output layers are language dependent while the rest of the layers are shared among the languages.
  \item Multilingual (5 languages): this architecture is similar to the previous one except it uses an additional fully connected layer of 1024 neurons, and two extra output layers corresponding to the 2 new languages.
\end{enumerate}
The increased number of layers is intended at modeling the extra training data gained by adding languages.

\section{Residual Networks}\label{sec:resnet}
\begin{figure}
 \centering
 \centerline{\includegraphics[width=\linewidth]{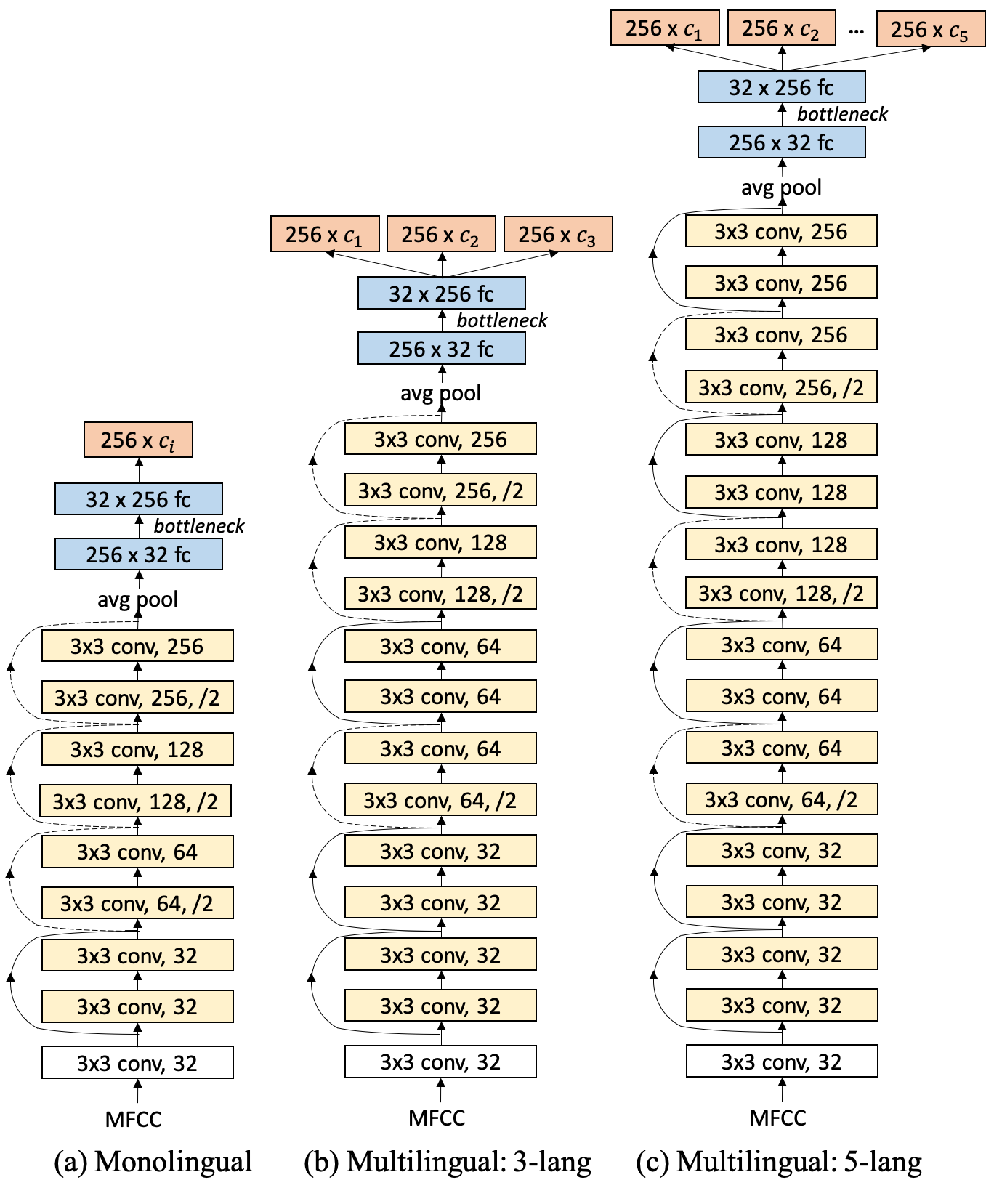}}
 \caption{Monolingual and multilingual residual network architectures for extracting bottleneck features using multiple languages. $c_i$ is the number of classes for the $i$-th language.}
 \label{fig:resnet}
\end{figure}

A Residual Network~\cite{he2016deep} is a CNN with shortcut connections between its stacked layers. Skipping layers effectively simplifies the training and gives flexibility to the network. 
Given an input matrix $x$ and an output matrix $y$, it models the function  $y = f(x) + x$ in each stacked layer, where $f(.)$ represents two convolutional layers with a non-linearity in-between. In case the size of the output of $f$ does not match the size of  $x$, one linear convolutional layer is applied to $x$ (implemented using $1\times1$ convolutions) before the addition operation.  Finally, a non-linearity is applied to the summed output $y$.

Similar to FFNs, we implemented 3 different architectures depending on the number of languages used for training. Those architectures are shown in Figure~\ref{fig:resnet}. We use $3\times3$ filters for all convolution layers throughout the network. Every time we reduce the feature map size by half (using a stride of 2), we double the number of filters. Then we perform a global average pooling to obtain 256 dimensional vector. These vectors are passed through a fully connected linear bottleneck layer which feeds to another layer of size 256. This goes to a single or multiple output classes depending on type of network: monolingual or multilingual. Smaller number of layers are used here in comparison to~\cite{he2016deep} due to the limited amount of training data.

\section{Experimental Setup}\label{sec:exp-setup}
In this section, we describe the databases and the pre-processing steps to perform the experiments. Then, we present the details of training different neural networks.

\subsection{Databases} \label{sec:database}
\begin{description}[leftmargin=0.3cm]
 \item[GlobalPhone Corpus:] 
GlobalPhone~\cite{schultz2013globalphone} is a multilingual speech database consisting of high quality recordings of read speech with corresponding transcription and pronunciation dictionaries in 20 different languages. 
In this work, we use French (FR), German (GE), Portuguese (PT), Spanish (ES) and Russian (RU) to train monolingual as well as multilingual networks and estimate the corresponding bottleneck features for QbE-STD experiments. 
These languages were chosen to have a complete mismatch between the training and test languages. 
We have an average of $\sim$20 hours of training and $\sim$2 hours of development data per language.
\item[Query by Example Search on Speech Task (QUESST):]
QUESST dataset~\cite{anguera2014query} is part of MediaEval 2014 benchmarking
initiative and is used here to evaluate the performance of different bottleneck features for QbE-STD. It consists of $\sim$23 hours of audio recordings (12492 files) in 6 languages as search corpus: Albanian, Basque, Czech, non-native English, Romanian and Slovak. The development and evaluation set includes 560 and 555 queries respectively which were separately recorded than the search corpus. The development queries are used to tune the hyperparameters of different systems. There are three types of occurrences of a query defined as a match in this dataset. Type 1: exactly matching the lexical representation of a query, Type 2: slight lexical variations at the start or end of a query, Type 3: multiword query occurrence with different order or filler content between words. (See~\cite{anguera2014query} for more details)
\end{description}

\subsection{Neural Networks Training} \label{sec:nn-train}
We use mel frequency cepstral coefficients (MFCC) with corresponding $\Delta$ and $\Delta\Delta$ features as input to the neural networks. The outputs are mono-phone based tied states (also known as pdfs in Kaldi~\cite{povey2011kaldi}) corresponding to each language as presented in Section~\ref{sec:database}. The training labels for these networks are generated using GMM-HMM based speech recognizers~\cite{hinton2012deep, tong2017investigation}. The number of classes corresponding to French, German, Portuguese, Spanish and Russian are 124, 133, 145, 130, 151 respectively. Note that, we also trained these networks using tri-phone based senone classes, however they perform worse than the mono-phone based training. All neural network architectures in this work is implemented using Pytorch~\cite{paszke2017pytorch}.

\begin{description}[leftmargin=0.3cm]
\item[Feed Forward Networks:] 
The input to the FFNs is MFCC features with a context of 6 frames (both left and right) resulting in a 507 dimensional vector. 
We apply layer normalization~\cite{ba2016layer} before the linear transforms and use rectifier linear unit (ReLU) as non-linearity after each linear transform except in the bottleneck layer.
We train those networks with batch size of 255 samples and dropout of 0.1. In case of multilingual training, we use equal number of samples from each language under consideration. Adam optimization algorithm~\cite{kingma2014adam} is used with an initial learning rate of $10^{-3}$ to train all networks by optimizing cross entropy loss. The learning rate is halved every time the development set loss increases compared to the previous epoch until a value of $10^{-4}$. All the networks were trained for 50 epochs. 

\item[Residual Networks:]
We construct the input for ResNet training using MFCC features with a context of 12 frames (both left and right) resulting in a $39\times25$ size matrix with single channel in contrast to the 3 channel RGB images generally used in image classification tasks. We also conducted experiments by arranging the input MFCC features in 3 channels: static, $\Delta$ and $\Delta\Delta$ values~\cite{abdel2014convolutional}, however the performance was worse. Batch normalization~\cite{ioffe2015batch} is applied after every convolution layer and ReLU is used as non-linearity. The networks are trained with batch size of 255 samples and dropout of 0.05 for 50 epochs. We use the same learning rate schedule as the FFNs with initial and final learning rate of $10^{-3}$ and $10^{-4}$ respectively.
\end{description}
The number of layers for both FFN and ResNet architectures for different monolingual and multilingual networks are optimized using the development queries to give best QbE-STD performance. The input context size for these networks are optimized as well by varying from it from 4 to 14. We observed the optimal context size corresponding to FFN and ResNet are 6 and 12 respectively. The performance gain of the ResNet models over FFN models (as we will see in Section~\ref{sec:exp}) indicates that ResNets are better equipped to capture information from longer temporal context than the FFNs.

\subsection{DTW for Template Matching}\label{sec:dtw}
The trained neural networks are used to estimate bottleneck features for DTW. As a pre-processing step, we implement a speech activity detector (SAD) by utilizing silence and noise class posterior probabilities obtained from three different phone recognizers (Czech, Hungarian and Russian)~\cite{schwarz2009phoneme} trained on SpeechDAT(E) database~\cite{pollak2000speechdat}. Those posterior probabilities are averaged and compared with rest of the phone class probabilities to find and remove the noisy frames. Any audio file with less than 10 frames after SAD is not considered for experiments. 

The DTW system presented in~\cite{rodriguez2014high} is used here to compute the matching score for a query and audio document pair. It utilizes cosine similarity to obtain the frame-level distance matrix from a query and an audio document. This DTW algorithm is similar to slope-constrained DTW~\cite{hazen2009query} where the optimal warping path is normalized by its partial path length at each step and constraints are imposed so that the warping path can start and end at any point in the audio document. The scores generated by the DTW system are normalized to have zero-mean and unit-variance per query in order to reduce variability across different queries~\cite{rodriguez2014high}.

\subsection{Evaluation Metrics} \label{sec:eval-metric}
Minimum normalized cross entropy ($C_{nxe}^{\min}$) is used as primary metric and maximum term weighted value ($MTWV$) is used as secondary metric to compare performances of different bottleneck features for QbE-STD~\cite{rodriguez2013mediaeval}. The costs of false alarm ($C_{fa}$) and missed detection ($C_m$) for $MTWV$ are considered to be 1 and 100 respectively. One-tailed paired-samples t-test is conducted to evaluate the significance of performance improvement. Additionally, detection error trade-off (DET) curves are used to compare the detection performance of different systems for a given range of false alarm probabilities. 

\begin{table*}[t]
\caption{Performance of the QbE-STD system in QUESST 2014 database using various monolingual and multilingual bottleneck features for different types of evaluation queries. $C_{nxe}^{\min}$ (lower is better) and $MTWV$ (higher is better) is used as evaluation metric.} \label{table:quesst}
 \begin{center}
 \begin{tabular}{|c|c|c|c|c|c|c|c|c|c|}
  \hline
  \parbox[t]{2mm}{\multirow{12}{*}{\rotatebox[origin=c]{90}{Monolingual}}}
  \parbox[t]{2mm}{\multirow{12}{*}{\rotatebox[origin=c]{90}{Feature}}}
  & Training &\multirow{2}{*}{System} & \multicolumn{2}{c|}{T1 Queries} & \multicolumn{2}{c|}{T2 Queries} & \multicolumn{2}{c|}{T3 Queries} \\
  & Language & & $C_{nxe}^{\min} \downarrow$ & $MTWV \uparrow$ & $C_{nxe}^{\min} \downarrow$ & $MTWV \uparrow$ & $C_{nxe}^{\min} \downarrow$ & $MTWV \uparrow$ \\
  \cline{2-9} \cline{2-9}

  & \multirow{2}{*}{Portuguese (PT)} & FFN & 0.5582 & 0.4671 & 0.6814 & 0.3048 & 0.8062 & 0.1915 \\
  & & ResNet & 0.5405 & 0.4698 & 0.6607 & 0.2747 & 0.7954 & 0.1802 \\
  \cline{2-9}
  & \multirow{2}{*}{Spanish (ES)} & FFN & 0.5788 & 0.4648 & 0.7074 & 0.2695 & 0.8361 & 0.1612 \\
  & & ResNet & 0.5718 & 0.4465 & 0.7043 & 0.2613 & 0.8465 & 0.1462 \\
  \cline{2-9}
  & \multirow{2}{*}{Russian (RU)} & FFN & 0.6119 & 0.4148 & 0.7285 & 0.2434 & 0.8499 & 0.1385 \\
  & & ResNet & 0.5728 & 0.4405 & 0.7017 & 0.2481 & 0.8525 & 0.1346  \\
  \cline{2-9}
  & \multirow{2}{*}{French (FR)} & FFN & 0.6266 & 0.4242 & 0.7462 & 0.2086 & 0.8522 & 0.1249 \\
  & & ResNet & 0.5957 & 0.4225 & 0.7017 & 0.2216 & 0.8540 & 0.1267 \\
  \cline{2-9}
  & \multirow{2}{*}{German (GE)} & FFN & 0.6655 & 0.3481 & 0.7786 & 0.1902 & 0.8533 & 0.1038 \\
  & & ResNet & 0.6389 & 0.3803 & 0.7511 & 0.2230 & 0.8497 & 0.1166 \\
  \hline \hline

  \parbox[t]{2mm}{\multirow{4}{*}{\rotatebox[origin=c]{90}{Concat.}}}
  \parbox[t]{2mm}{\multirow{4}{*}{\rotatebox[origin=c]{90}{Feature}}}
  & \multirow{2}{*}{PT-ES-RU} & FFN & 0.5450 & 0.4957 & 0.6665 & 0.2985 & 0.8053 & 0.1869 \\
  & & ResNet & 0.5072 & 0.5164 & 0.6374 & 0.3162 & 0.7965 & 0.1899 \\
  \cline{2-9}
  & \multirow{2}{*}{PT-ES-RU-FR-GE} & FFN & 0.5457 & 0.4965 & 0.6715 & 0.2903 & 0.8079 & 0.1930 \\
  & & ResNet & 0.5040 & 0.5201 & 0.6309 & 0.3212 & 0.7941 & 0.1914 \\
  \hline \hline

  \parbox[t]{2mm}{\multirow{4}{*}{\rotatebox[origin=c]{90}{Multiling.}}}
  \parbox[t]{2mm}{\multirow{4}{*}{\rotatebox[origin=c]{90}{Feature}}}
  & \multirow{2}{*}{PT-ES-RU} & FFN & 0.4828 & 0.5459 & 0.6218 & 0.3626 & 0.7849 & 0.2057 \\
  & & ResNet & 0.4554 & 0.5666 & 0.6009 & 0.3529 & 0.7650 & 0.2201 \\
  \cline{2-9}
  & \multirow{2}{*}{PT-ES-RU-FR-GE} & FFN & 0.4606 & 0.5663 & 0.6013 & 0.3605 & 0.7601 & 0.2138 \\
  & & ResNet & {\bf 0.4345} & {\bf 0.5962} & {\bf 0.5703} & {\bf 0.3815} & {\bf 0.7387} & {\bf 0.2487} \\
  \hline
 \end{tabular}
\vspace{-3mm}
 \end{center}
\end{table*}

\begin{table}
\caption{Number of parameters for different mono and multi lingual models using FFN and ResNet architecture.} \label{table:parameters}
 \begin{center}
 \begin{tabular}{|c|c|c|}
  \hline
  Model & FFN & ResNet \\
  \hline \hline
  Monolingual & $\sim$ 1.8M  & $\sim$ 663K \\
  Multilingual: 3-lang & $\sim$ 3.1M & $\sim$ 1.4M \\
  Multilingual: 5-lang & $\sim$ 4.4M & $\sim$ 3.0 M \\
  \hline
 \end{tabular}
 \end{center}
\end{table}

\section{Experimental Analysis}\label{sec:exp}
In this section, we report and analyze the QbE-STD performance using various bottleneck features estimated from our FFN and ResNet models. Previously, the best performance on QUESST 2014 database was obtained using monolingual bottleneck features estimated using FFNs~\cite{szoke2015copingwith}. We implemented those models to compare with multilingual features as well as corresponding ResNet based models.

\subsection{Monolingual Feature Performance} \label{sec:monoperf}
We train five different monolingual networks for both architectures: FFN and ResNet, corresponding to PT, ES, RU, FR, GE languages from GlobalPhone database. We evaluate the features estimated with these networks using QbE-STD as described in Section~\ref{sec:dtw}. Similar to~\cite{szoke2015copingwith}, we did not employ any specific strategies to deal with different types of queries in QUESST 2014. 
The results are presented using $C_{nxe}^{\min}$ and $MTWV$ metrics in Table~\ref{table:quesst}. We can see that the ResNet based bottleneck features perform better than most of the FFN based features in terms of $C_{nxe}^{\min}$ metric, except for T3 queries with FR, ES and RU features, where the performances are close. We also observe that PT features perform best for both FFN and ResNet.

\subsection{Multilingual Feature Performance} \label{sec:multiperf}
We present the results of our multitask learning based multilingual systems and compare their performance with a simple monolingual feature concatenation approach.
\begin{description}[leftmargin=0.3cm]
\item[Multitask Learning:] 
We implement two multilingual networks corresponding to each FFN and ResNet architectures discussed in Sections~\ref{sec:ffn} and~\ref{sec:resnet} using 3 languages (PT, ES, RU) and 5 languages (PT, ES, RU, FR, GE). The 3 language network uses the best performing monolingual training languages. Performance of the features extracted from these networks are shown in Table~\ref{table:quesst}. Clearly, ResNet based bottleneck features provide significant improvement over the corresponding FFN based features. We also observe that PT-ES-RU-FR-GE features significantly outperform PT-ES-RU features for both FFN and ResNet model indicating that additional languages for training provide better language independent features.


\item[Feature Concatenation:]
Another way of utilizing training resources from multiple languages is to concatenate the monolingual bottleneck features to perform DTW. We perform two sets of experiments by concatenating monolingual features from PT-ES-RU and FR-GE-PT-ES-RU languages corresponding to both FFN and ResNet. The results are presented in Table~\ref{table:quesst}. We can see that there is marginal improvement over the best monolingual feature (PT) from FFN model, a similar observation was presented in~\cite{szoke2015copingwith}. On the other hand, ResNet based feature (PT-ES-RU) perform significantly better than the corresponding PT features. However, there is no significant performance difference between the ResNet based 3 and 5 language feature concatenation.
\end{description}

We also observe that the multitask learning based features significantly outperform the monolingual feature concatenation, indicating the importance of multitask learning for utilizing training resources from multiple languages.

The number of parameters for different models are shown in Table~\ref{table:parameters}. We observe that the FFNs have more parameters than the ResNet architectures. The improved performance of ResNet models in comparison to FFNs indicate that ResNet architecture produces better bottleneck features in spite of having less parameters.

\subsection{Monolingual vs Multilingual Feature}
\begin{figure}
  \centering
  \vspace{-10mm}
  \centerline{\includegraphics[width=\linewidth]{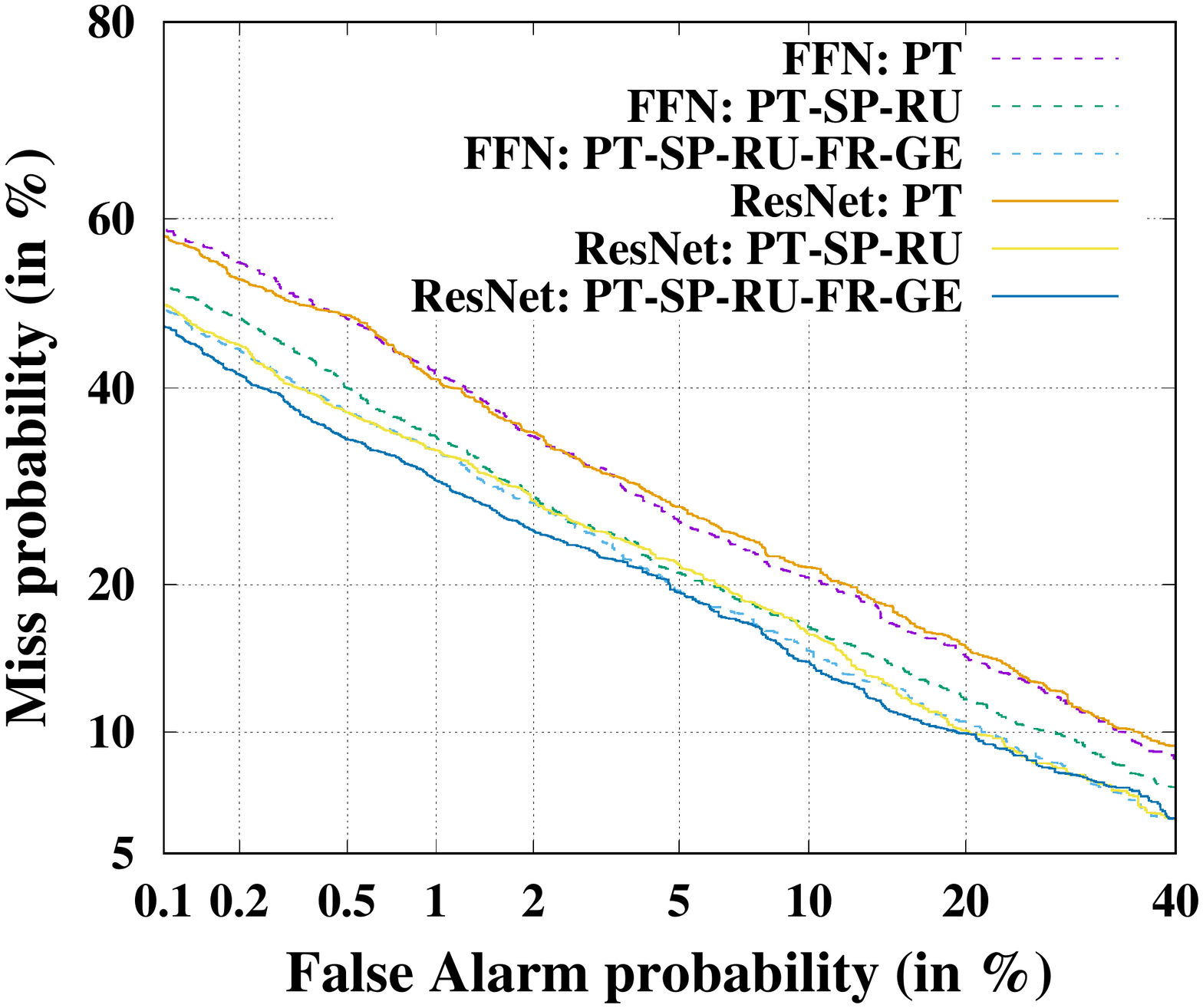}}
  \vspace{-5mm}
  \caption{DET curves showing the performance of monolingual and multilingual features estimated using FFNs and ResNets for T1 queries of QUESST 2014.}
\label{fig:det}
\vspace{-6mm}
\end{figure}
\begin{figure}
  \centering
  \centerline{\includegraphics[width=\linewidth]{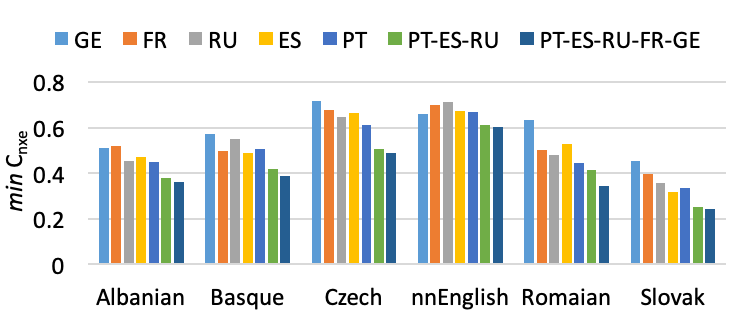}}
  \vspace{-3mm}
  \caption{Comparison of QbE-STD performance of language specific evaluation queries (T1 query) using $C_{nxe}^{\min}$ values} 
\label{fig:mcnxe}
\vspace{-5mm}
\end{figure}
The 3 language multilingual feature provides an average absolute gain of 5.2\% and 5.8\% (in $C_{nxe}^{\min}$) for FFN and ResNet model respectively in comparison to the corresponding best monolingual features. Further 2.3\% and 2.5\% absolute improvements are observed while using 2 more languages for training.
In order to compare the missed detection rate for a given range of false rates we present the DET curves corresponding to these systems in Figure~\ref{fig:det}. We see a similar trend of performance improvement here as well. We also observe that the performance gain is higher from 1 language to 3 languages than 3 languages to 5 languages. It is due to our use of the best performing languages to train the 3 language network.


\subsection{Language Specific Performance}\label{sec:lang-spe}
We compare the language specific query performance of ResNet based monolingual and multilingual features as it performs better than the FFN counterparts. We use $C_{nxe}^{\min}$ values of T1 query performance to show this comparison in Figure~\ref{fig:mcnxe}. We observe that the performance improves with more languages used for training, however the amount of improvement varies with language of the query. The smaller performance gain from 3 to 5 languages for some queries (e.g. Albanian, Czech, Slovak) can be attributed to much worse performance of FR and GE features compared to rest of the monolingual features.


\section{Conclusions}\label{sec:con}
We proposed a ResNet based neural network architecture to estimate monolingual as well as multilingual bottleneck features for QbE-STD. We present a performance analysis of these features using both ResNets and FFNs. It shows that additional languages for training improves performance and the ResNets perform better than FFNs for both monolingual and multilingual features. Further analysis shows that the improvement is consistent throughout queries of different languages. In future, we plan to train deeper ResNets with more languages to compute and analyze language independence of those features. The improved bottleneck features can be used for other relevant tasks e.g. unsupervised unit discovery.


\bibliographystyle{IEEEbib}


\bibliography{phaser,refs,bibliography}

\end{document}